\documentclass[final]{cvpr}

\usepackage{times}
\usepackage{epsfig}
\usepackage{graphicx}
\usepackage{amsmath}
\usepackage{amssymb}
\usepackage{array}
\usepackage{multirow}

\usepackage[pagebackref=true,breaklinks=true,colorlinks,bookmarks=false]{hyperref}

\makeatletter
\def\hlinewd#1{%
\noalign{\ifnum0=`}\fi\hrule \@height #1 \futurelet
\reserved@a\@xhline}
\newcommand{\hthickline}{\hlinewd{1.0pt}}

\makeatother

\begin{document}

\title{FRDet: Balanced and Lightweight Object Detector based on Fire-Residual Modules for Embedded Processor of Autonomous Driving}

\author{Seontaek Oh, Ji-Hwan You, Young-Keun Kim\\ School of Mechanical and Control Engineering, Handong Global University\\
Pohang, Rep. Korea\\
{\tt\small ykkim@handong.edu}
}

\maketitle

\begin{abstract}
For deployment on an embedded processor for autonomous driving, the object detection network should satisfy all of the accuracy, real-time inference, and light model size requirements.
Conventional deep CNN-based detectors aim for high accuracy, making their model size heavy for an embedded system with limited memory space. In contrast, lightweight object detectors are greatly compressed but at a significant sacrifice of accuracy. 
Therefore, we propose FRDet, a lightweight one-stage object detector that is balanced to satisfy all the constraints of accuracy, model size, and real-time processing on an embedded GPU processor for autonomous driving applications.
Our network aims to maximize the compression of the model while achieving or surpassing YOLOv3’s level of accuracy.
This paper proposes the Fire-Residual (FR) module to design a lightweight network with low accuracy loss by adapting fire modules with residual skip connections. In addition, the Gaussian uncertainty modeling of the bounding box is applied to further enhance the localization accuracy. 
Experiments on the KITTI dataset showed that FRDet reduced the memory size by 50.8\% but achieved higher accuracy by 1.12\% mAP compared to YOLOv3. Moreover, the real-time detection speed reached 31.3 FPS on an embedded GPU board(NVIDIA Xavier).
The proposed network achieved higher compression with comparable accuracy compared to other deep CNN object detectors while showing improved accuracy than the  lightweight detector baselines.
Therefore, the proposed FRDet is a well-balanced and efficient object detector for practical application in autonomous driving that can satisfies all the criteria of accuracy, real-time inference, and light model size.

\end{abstract}

\section{Introduction}

Recently, there has been significant progress in the accuracy and inference speed of image-based object detection for autonomous vehicles. For practical  deployment in an autonomous vehicle, the object detection network should be optimized to satisfy the following conditions. First, for safe driving, the object detector should meet the high accuracy requirement. Second, it should be a lightweight model for deployment on an embedded processor in the autonomous vehicle. 
Third, it should have a real-time inference capability for low latency in control and communication even on the embedded platform. 
The inference time is highly correlated with the small size of the detector model, but there is a trade-off between inference time and accuracy. Maximizing the model size compression while minimizing the accuracy loss is an ideal condition for designing an object detector for an actual application on autonomous vehicles.

The approaches to image object detectors based on deep learning are classified into either the two-stage or the one-stage categories. The two-stage methods, such as R-CNN~\cite{girshick2014rich}, Fast R-CNN~\cite{girshick2015fast}, and Faster R-CNN~\cite{ren2015faster}, detect objects through a series of stages using a region proposal network and classification network. Object detectors based on this approach have achieved state-of-the-art performance for accuracy but have a lower inference speed. In contrast, the one-stage methods, such as YOLO~\cite{redmon2016you} and SSD~\cite{liu2016ssd}, demonstrate a higher inference speed but show relatively lower accuracy than the two-stage methods. 

In terms of real-time processing, one-stage detectors are preferable for autonomous driving. Recently, there have been some notable achievements for one-stage detectors on improving the detection accuracy and compressing the model size for faster and lighter networks. 

Advancements in the accuracy of one-stage detectors were made possible by applying various techniques, which are described in Section~\ref{ref_objectdetection}. For example, YOLOv2~\cite{redmon2017yolo9000} and YOLOv3~\cite{redmon2018yolov3} can perform faster and more robustly than YOLOv1~\cite{redmon2016you} by applying the latest methods, including deeper networks with residual structures, batch normalization, and pyramid scales. RefineDet~\cite{zhang2018single} improved the accuracy as a single-shot detector by introducing an anchor refinement module. RetinaNet~\cite{lin2017focal} surpasses the accuracy of two-stage detectors by applying focal loss to minimize the foreground-background class imbalance. Gaussian YOLOv3~\cite{choi2019gaussian} improves the detection accuracy over YOLOv3 by predicting the localization uncertainty of the boundary box with a Gaussian parameter. These improved object detectors attain high accuracy and real-time inference by running on a high-performance GPU platform with abundant memory. However, their model sizes are too large for embedded processors with limited memory space. 

With the advent of embedded GPUs and EdgeAI, compression of the detector model while minimizing the loss of accuracy has gained increased attention. SqueezeDet~\cite{wu2017squeezedet}, a detector based on the fire modules of SqueezeNet~\cite{iandola2016squeezenet}, could reduce the model size to 30 times smaller than previous detector baselines. MobileNet v1$\sim$v3~\cite{howard2017mobilenets, sandler2018mobilenetv2, howard2019searching} significantly reduced the model parameters using the concept of depthwise separable convolution, pointwise convolution, and inverted residuals with linear bottlenecks. Furthermore, PeleeNet~\cite{wang2018pelee}, Tinier-YOLO~\cite{fang2019tinier}, Mini-YOLOv3~\cite{mao2019mini}, and YOLO-LITE~\cite{huang2018yolo} were proposed as a very light model for real-time inference on mobile platform. Although these lightweight compressed models are designed for real-time inference on an embedded  platform, there is a high loss of accuracy due to the trade-off between accuracy and speed.

Therefore, this paper addresses this trade-off issue by proposing an efficient and balanced object detection network. Instead of focusing the design weight primarily on either accuracy or memory size, we distributed more balanced weight on the accuracy, model size and the speed criteria for autonomous driving applications.

We propose FRDet, a fire-residual module based detector, that efficiently compresses the memory size for real-time deployment on an embedded processor while achieving YOLOv3 level accuracy.
This study designed the Fire-Residual (FR) module that combines residual skip connection~\cite{he2016deep} with the fire module from  SqueezeNet~\cite{iandola2016squeezenet} to compress the model’s weighting parameters effectively while minimizing accuracy loss. Based on the FR modules, the FRDet model was designed with an architecture inspired by YOLOv3~\cite{redmon2018yolov3}.
The pipeline of the proposed detector starts by extracting the feature maps through the proposed FR module. Then, the feature maps are transferred to the multi-scale detection layers of feature pyramid networks ~\cite{lin2017feature} for accurate scale-aware detection. Further to improve the boundary box regression with a reduced memory burden, the localization of the boundary box was based on Gaussian parameters as in Gaussian YOLOv3~\cite{choi2019gaussian}.

The performance of FRDet was investigated using KITTI datasets~\cite{geiger2012we} to validate the deployment on an embedded processor for autonomous driving.
The accuracy, memory size, and inference speed were evaluated on the embedded GPU processor of NVIDIA Xavier.
The test results showed that the proposed detector reduced the memory size by 50.8\% but achieved 1.12\% higher accuracy (mAP) compared to YOLOv3. The inference detection speed on the embedded processor could achieve 31.3 FPS with an image resolution of 480x360. 
Moreover, compared to previous deep CNN and lightweight object detectors, FRDet exhibited the advantage of having a balanced accuracy, memory size, and speed for embedded processors with limited memory. 

The remainder of this paper is organized as follows. In Section 2, we review related works on object detectors with improved accuracy and compressed model size. In Section 3, the FRDet pipeline is introduced with the description of FR modules. Evaluation of the accuracy, memory size, and speed on the embedded GPU processor that were performed on KITTI datasets are analyzed in Section 4. Then, Section 5 concludes the paper.



\section{Related Works} \label{ref_objectdetection}

\subsection{Object Detection on Deep CNN} 
\vspace{0.3\baselineskip}
\textbf{Two-Stage Approach.}
Deep CNN-based object detection can be classified into two categories: either the two-stage or one-stage approaches. Early CNN object detectors predominantly used the two-stage detector method. The two-stage approach detects objects through the stages of generating sparse region proposal sets and classifying them. Notably, R-CNN~\cite{girshick2014rich} was one of the first two-stage approaches that used a selective search for generating proposals that are fed into deep CNN-based classification and bounding box regression. R-CNN was improved by the introduction of Fast R-CNN~\cite{girshick2015fast} and Faster R-CNN~\cite{ren2015faster} that achieved a higher inference speed and the best accuracy on various challenging datasets such as PASCAL VOC 2012~\cite{everingham2010pascal}, MS COCO ~\cite{lin2014microsoft}, and KITTI ~\cite{geiger2012we}. However, their detection speed could not meet the real-time inference of the one-stage approach. There have been numerous developments to improve the inference speed of the two-stage approach.
MS-CNN~\cite{cai2016unified} improved the inference speed by reducing the memory with implementation of feature upsampling by deconvolution. 
MaxpoolNMS\cite{cai2019maxpoolnms} accelerated region proposal networks process with parallel processing of max-pooling classification score maps.

\vspace{0.3\baselineskip}
\textbf{One-Stage Approach.} Real-time detection inference was possible with the introduction of a one-stage approach, notably SSD~\cite{liu2016ssd} and YOLO~\cite{redmon2016you}. One-stage detection combines multi-labeled classification and bounding box regression simultaneously in a single feed-forward CNN. SSD~\cite{liu2016ssd} created multiple feature maps with a set of default boxes with different aspect ratios and scales for scale-aware detection. YOLO~\cite{redmon2016you} divided the input image into grids and predicted bounding box coordinates and class probability for bounding boxes at each grid in a single CNN pipeline. YOLO achieved one of the fastest inference speeds compared to conventional detectors but produced higher detection errors in small and dense objects. To address the above issues and improve the accuracy, YOLOv2~\cite{redmon2017yolo9000} adopted batch normalization, anchors, and multi-scale training with a deeper CNN. YOLOv3~\cite{redmon2018yolov3} improved over YOLOv2 by adopting residual skip connections and the Feature Pyramid Network (FPN)~\cite{lin2017feature} architecture that extracts multi-scale feature maps to improve the detection performance for scale-invariant detection. 

Although the one-stage detectors could achieve fast inference, they had relatively lower accuracy than the two-stage methods. Recent studies focused on pursuing high accuracy for the one-stage approach with various techniques~\cite{lin2017focal, zhang2018single, choi2019gaussian}. RetinaNet~\cite{lin2017focal} proposed a new loss function called focal loss that effectively deals with the foreground-background class imbalance and may increase detection accuracy. RefineDet~\cite{zhang2018single} further improved the regression accuracy by the anchor refinement module that filters out negative anchors and adjusts the anchor locations and sizes. Gaussian YOLOv3~\cite{choi2019gaussian} replaced the bounding box coordinates with Gaussian parameters to predict the localization uncertainty, which could reduce mislocalization errors. 

One-stage detectors with improved accuracy have attracted significant attention for autonomous driving applications. However, their real-time evaluation was based on powerful GPUs, and their model size was large for deployment on an embedded platform with limited memory space. 

\subsection{Object Detection on Compressed CNN} 
\vspace{0.3\baselineskip}
\textbf{Compressed CNN Model.}
Numerous lightweight CNN models with comparable high accuracy for mobile applications have been proposed in recent years~\cite{iandola2016squeezenet, huang2017densely, huang2018condensenet, wang2018pelee, howard2017mobilenets, sandler2018mobilenetv2, ma2018shufflenet, zhang2018shufflenet}.
SqueezeNet~\cite{iandola2016squeezenet} proposed a lightweight DNN architecture with the fire module that squeezes input channels with 1x1 kernels and expands the output with 3x3 and 1x1 kernels, which consequently reduces the weight parameters while minimizing accuracy loss. SqueezeNet achieved AlexNet~\cite{krizhevsky2012imagenet} level accuracy while compressing the architecture to less than 0.5 MB. MobileNet~\cite{howard2017mobilenets} introduced the concept of depthwise separable convolution to design an efficient model for mobile devices ~\cite{sifre2014rigid}. The depthwise separable convolution replaced the standard convolution by the spatial depthwise convolution followed by the channel pointwise convolution. MobileNet v2~\cite{sandler2018mobilenetv2} further improved both the speed and memory usage by applying inverted residuals and linear bottlenecks. It could run real-time inference even on a CPU platform. ShuffleNet~\cite{ma2018shufflenet,zhang2018shufflenet} also used depthwise convolution but added pointwise group convolution with channel shuffle to design a lightweight CNN with higher accuracy than MobileNet v1. It achieved 13x faster speed over AlexNet on an embedded CPU device. DenseNet~\cite{huang2017densely} proposed a dense CNN that could reduce the number of direct layer connections greatly by enabling each layer to receive from all the preceding layers. The reduction in the connections with a fewer number of channels could make the network with higher memory efficiency. CondenseNet~\cite{huang2018condensenet} adopted the learned group convolution module on DenseNet to remove the excessively dense connections thereby using only 10\% of the computation load with accuracy comparable to DenseNet. PeleeNet~\cite{wang2018pelee} is another efficient architecture for mobile devices based on DenseNet and
achieved higher accuracy and faster inference at a smaller size compared to MobileNet v2. 

\vspace{0.3\baselineskip}
\textbf{Compressed Object Detector}
By combining the aforementioned compressed CNN methods mainly on SSD and YOLO detection architecture, efficient lightweight object detection has been developed. SSDLite~\cite{sandler2018mobilenetv2} is a mobile-applicable object detector that implemented the depthwise separable convolution of MobileNetv2 on the SSD architecture. Pelee~\cite{wang2018pelee} applied the PeleeNet model with a residual prediction block and a small convolutional kernel prediction on the SSD architecture. It achieved 125 FPS on an embedded GPU with 66\% of the model size of MobileNetv1. Tiny-SSD~\cite{womg2018tiny} adopted SqueezeNet fire modules on SSD for high-speed inference.

Tiny-YOLOv3~\cite{redmon2018yolov3} is a lightweight version of YOLOv3 with a smaller model size that can run in real-time on embedded GPUs. 
YOLO-LITE~\cite{huang2018yolo}, based on YOLOv2, is a lighter and faster model than SSDLite and Tiny-YOLOv3, which can run at 21 FPS on a non-GPU.
Mini-YOLOv3~\cite{mao2019mini} is another lightweight detector based on YOLOv3, which reduced the number of parameters by approximately 77\% with the application of depth-separable convolutions and pointwise group convolutions. 
SqueezeDet~\cite{wu2017squeezedet} and Tinier-YOLO~\cite{fang2019tinier} proposed energy-efficient and high-speed object detectors by using SqueezeNet fire modules on YOLO architectures. 
SqueezeDet is one of the most robust lightweight models for autonomous driving applications and achieved an accuracy of as much as 80.4\% with a model size of 8 MB. 
Tinier-YOLO is only 9 MB in size but the accuracy is reduced to about 60\% on MS-COCO datasets, compared to YOLOv3.

These compressed object detectors have dramatically reduced their model sizes for embedded processor applications but at the expense of accuracy. 
For safe driving, the accuracy of the lightweight object detectors must be improved further. 

Therefore, we propose an object detector with more balanced weighting for the accuracy, model size, and speed criteria for practical applications in autonomous driving. The purpose of this paper is to design a one-stage detector that can match or surpass the accuracy of YOLOv3 while reducing the model size by approximately 50\% and running in real-time on an embedded GPU.

\begin{figure*} 
\begin{center}
\includegraphics[width=0.9\linewidth]{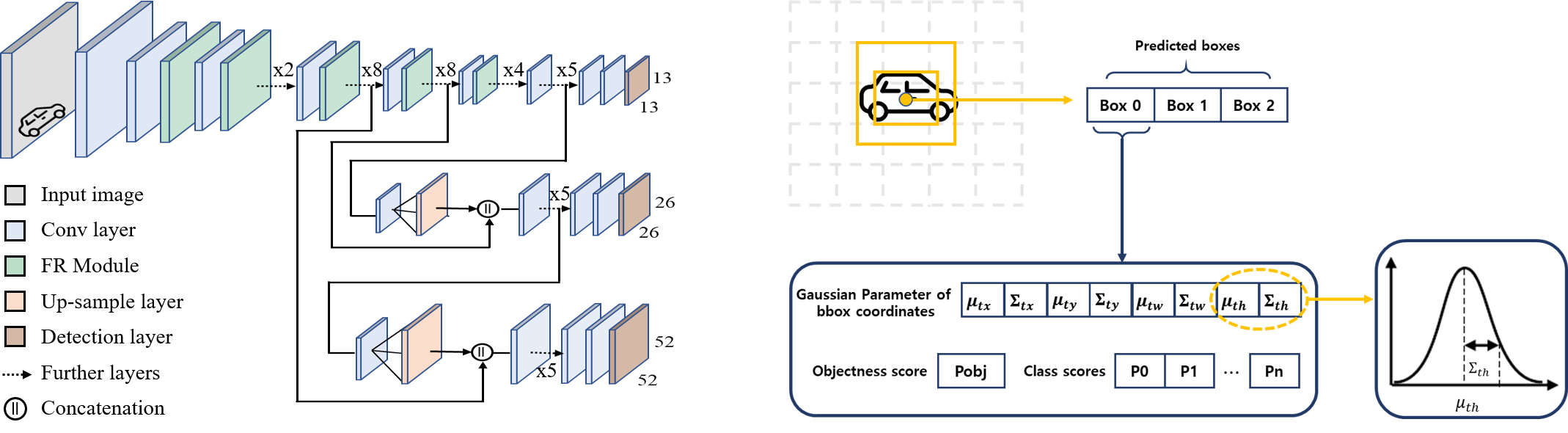}
\end{center}
\caption{Overview of the proposed FRDet. It is inspired by the YOLOv3~\cite{redmon2018yolov3} architecture and is composed of the proposed Fire-Residual(FR) modules and detection layers with Gausian parameters.The left panel indicates the pipeline of FRDet. The right describes the detection layer with Gaussian parameters. 
}
\label{fig:architecture}
\end{figure*} 

\begin{figure*} 
\begin{center}
\includegraphics[width=0.9\linewidth]{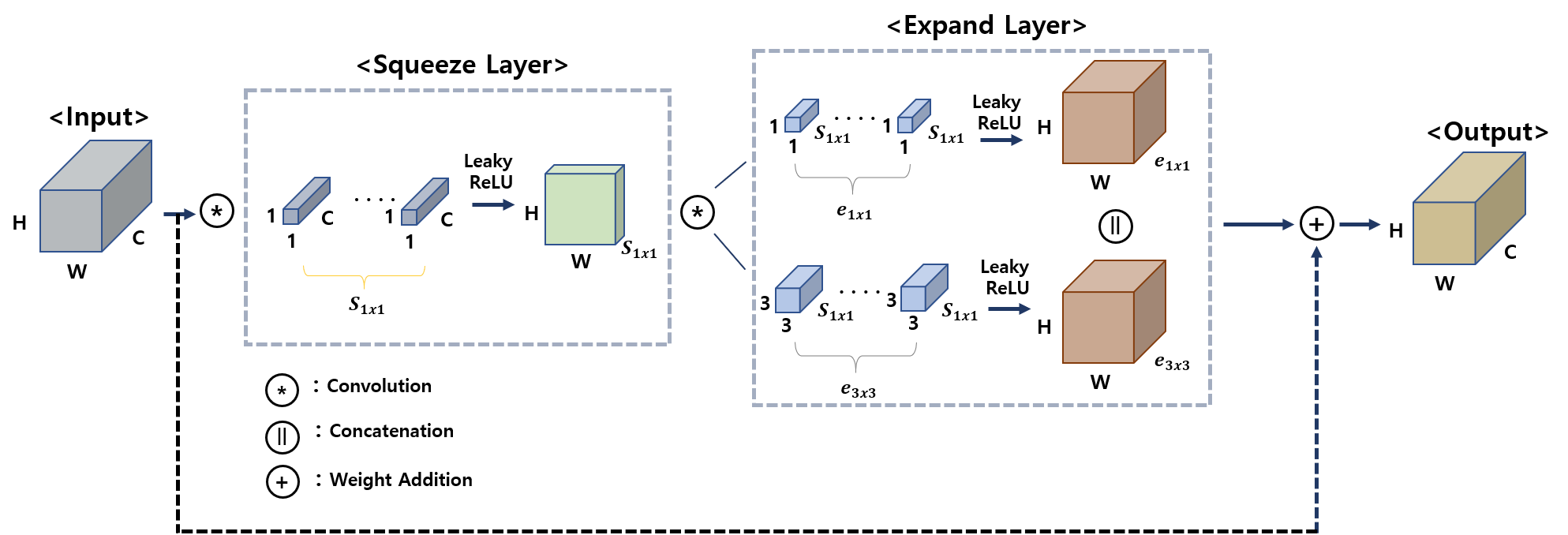}
\end{center}
\caption{Overview of the proposed FR module that is composed of three parts: the squeeze layer, expand layer, and residual skip connection.}
\label{fig:frmodule}
\end{figure*} 

\section{FRDet Architecture}
The overview of the proposed FRDet architecture is described in Figure~\ref{fig:architecture}.
The FRDet pipeline is inspired by the YOLOv3~\cite{redmon2018yolov3} architecture and is composed of a backbone CNN network based on the proposed Fire-Residual (FR) modules and detection layers. The FR module is a combination of the fire module with residual connections for a smaller and faster network and is responsible for extracting multi-scale feature maps from an input image using the pyramidal hierarchy of the feature pyramid network (FPN)~\cite{lin2017feature}. We also used standard convolution layers with 3x3 kernels with batch normalization and the activation function of leaky ReLU to downsample the feature maps.
The multi-scale feature maps are fed into the detection layers to predict the bounding box information and object classification.

\subsection{FR(Fire-Residual) Module} \label{FRmodule_description}
We designed the Fire-Residual (FR) module adopted from SqueezeNet~\cite{iandola2016squeezenet} to reduce the model size for low memory and high speed. To complement the accuracy loss due to compression, a residual skip connection is added.
As shown in Figure~\ref{fig:frmodule}, the FR module is composed of three parts: the squeeze layer, expand layer, and residual skip connection. The squeeze layer decreases the memory size of the network by reducing the input channels for the expand layer by applying 1D convolution kernels. The expand layer generates the feature maps that are concatenated from the outputs of the 1x1 and 3x3 kernel convolutions. Then, the concatenated output from the expand layer is added with the input weighting with the residual skip connection to generate the final output of the module. 

\vspace{0.3\baselineskip}
\textbf{Squeeze Layer and Expand Layer:} 
The squeeze and expand layers greatly reduce the number of weighting parameters while maintaining a high level of accuracy.
The squeeze layer aims to compress the number of weighting parameters by using only 1x1 kernels, which results in 9x fewer parameters than using 3x3 kernels. This acts as a bottleneck layer that reduces the number of input channels to the following expand layers for fewer calculations.  
The output that results from applying $s_{1\times1}$ numbers of 1x1 kernels are stacked and passed to the expand layer that consists of two parallel convolutions of 1x1 and 3x3 kernels. The expanded feature map is then generated by concatenating the feature maps from each parallel convolution, as shown in Figure ~\ref{fig:frmodule}.

In the case of convolution with 3x3 kernels, the total parameters($P_c$) in the network are measured by the following equation with the number of input channels ($C$), kernel size($3^2$), and kernel number($N$):

\begin{equation}
     W_c=3^2 \times C \times N, ~ B_c=N   
\end{equation}
\begin{equation}
     P_c = W_c + B_c = 3^2 \times C \times N + N
\end{equation}
where $W_c$ and $B_c$ are the parameter numbers of weight and bias.

In contrast, the parameters ($P_f$) in the proposed FR module can be calculated using the following equation:

\begin{eqnarray}
 P_f=(1^2 \times C \times s_{1\times1} + s_{1\times1}) \nonumber\\+ (1^2 \times s_{1\times1} \times e_{1\times1} + e_{1\times1}) \nonumber\\+ (3^2 \times s_{1\times1} \times e_{3\times3} + e_{3\times3})
\end{eqnarray}
where $s_{1\times1}$ is the number of kernels in the squeeze layer, and $e_{1\times1}$ and $e_{3\times3}$ are the numbers of 1x1 and 3x3 kernels in the expand layer,respectively. 
According to the equation, the reduction in the number of parameters depends on the number of 1x1 kernels in the squeeze layers ($s_{1\times1}$). 

For this study, we use the ratio between the number of kernels in the squeeze layer ($s_{1\times1}$) and the expand layer ($e_{1\times1} + e_{3\times3}$) as 1:16 with the reason explained in Section 4.1. With this ratio, we could compressed the memory size to 50.8\% of that of YOLOv3.

\vspace{0.3\baselineskip}
\textbf{Residual Skip Connection:} 
The substitution of a standard convolution layer by the squeeze and expand layers is useful to deepen the network at a smaller model size, but this brings the possibility of accuracy degradation. Thus, residual skip connections~\cite{he2016deep} are applied in the FR module to prevent degradation and preserve accuracy.
The input is connected to the concatenated feature map from the expand layer for the weighting addition, as shown in Figure ~\ref{fig:frmodule}.

\vspace{0.3\baselineskip}
\textbf{FR module design strategies:} 
The design of the FR module followed the following three strategies. First, when the number of input channels is $C$, then the number of output channels of the concatenate feature map should also be the same $C$ to effect the weighting addition of the residual connection. Thus, the total number of 1x1 and 3x3 kernels in the expand layer should be $ e_{1 \times 1}+e_{3 \times 3}=C$. Second, the number of each 1x1 and 3x3 kernels of the expand layer are set to be equal ($e_{1\times1} = e_{3\times3}$). Finally, the number of the squeeze layer ($s_{1\times1}$) is set to be less than the number of input channels ($C$) to compress the model size with the relationship of $s_{1\times1,k} = \frac{C}{2^k}, k=1,2,..$.

The hyperparameter tuning of $s_{1\times1,k}$ for high memory efficiency is studied by correlating the accuracy and memory size due to the value of $k$. The results showed that the model showed high accuracy when the squeeze ratio($2^k$) was 8 and 16, ~\textit{i.e.}, $k = 3, 4$. The detailed hyperparameter tuning is explained in Section~\ref{ablation_study}. 


\subsection{Feature Pyramid Networks (FPN)}
FRDet also uses the Feature Pyramid Network (FPN)~\cite{lin2017feature} for scale-invariant detection to enhance feature extraction. 
In brief, the FPN is a convolution network with a top-down pathway and lateral connections that construct rich and multi-scale feature maps from a single input image. Each level of the pyramid can be used to detect objects at different scales.

For FRDet, five levels($l$) of scale pyramids and two lateral connections were included. Each level of the pyramids has $2^l$ lower resolutions and more channels than the input (size of 416x416) to form multi-scale feature maps using a single input image. Moreover, for high-quality feature maps, two lateral connections are used by concatenating two feature maps, one from the detection layer and the other from the backbone layers, as shown in Figure ~\ref{fig:architecture}.

\subsection{Detection Layers}
The multi-scale detection layers are inspired by YOLOv3, which predicts the bounding box parameters including the bounding box locations, objectness scores, and class probability scores. The bounding box coordinates locate the center coordinates ($t_x$ and $t_y$), height, and width ($t_h$ and $t_w$) of the predicted boxes. The objectness score is the probability of an object’s existence, and the class score is the probability of the class object classification. 

A detection layer is divided into $S \times S$ grids and predicts $N$ anchor boxes for each grid cell. In this study, we divided the detection layers into $13 \times 13$, $26 \times 26$, and $52 \times 52$ grids for the different scales and used three anchor boxes for each grid per scale. Therefore, the tensor for the detection layer at one scale should be $S \times S \times [3\times (4 + 1 + N_C)]$ for predicting the four bounding box coordinates, one objectness score, and $N_C$ class predictions. 

To improve further the detection accuracy without additional memory burden, we applied Gaussian modeled boundary box regression and the loss function proposed by Gaussian YOLOv3~\cite{choi2019gaussian}. 
In contrast to the deterministic regression of the boundary box coordinates in YOLOv3, the localization uncertainty of the box coordinates is modeled with Gaussian parameters of the mean and variance functions. As shown in Figure~\ref{fig:architecture}, the four boundary box coordinate parameters were replaced by the mean values of ($\mu_{tx}$, $\mu_{ty}$, $\mu_{th}$, $\mu_{tw}$), and the variance values of ($\Sigma_{tx}$, $\Sigma_{ty}$, $\Sigma_{th}$, $\Sigma_{tw}$).
The loss function of the boundary box was also modified to accommodate the Gaussian parameters. A negative log-likelihood (NLL) loss function~\cite{choi2019gaussian} is used, while the loss function for other parameters of objectness and class classification remain the same. For example, the NLL loss function for the $t_x$ coordinate is expressed as

\begin{eqnarray}
 L_x = -\sum_{i=1}^{W}\sum_{j=1}^{H}\sum_{k=1}^{K}{\gamma_{ijk}} log(N(x^G_{ijk}|\mu_{t_x}(x_{ijk}),\nonumber\\\Sigma_{t_x}(x_{ijk})) + \varepsilon,
\end{eqnarray}
where $W$ and $H$ are the numbers of grids of each width and height of the feature map, respectively, and $K$ is the number of anchors. The ground truth of the $t_x$ coordinate is $x^G_{ijk}$, $\mu_{t_x}(x_{ijk})$ and $\Sigma_{t_x}(x_{ijk})$ are the mean and variance, respectively, of the uncertainty for the $t_x$ coordinates at the k-th anchor in the (i,j) grid cell. Similarly, other loss functions for $t_y, t_w, t_h$ are the same but with different parameters accordingly.

\section{Experiment}
For the experiment, we used the KITTI dataset~\cite{geiger2012we} to evaluate the object detection model for autonomous driving applications. 
The KITTI dataset is commonly used for detecting the classes of cars, cyclists, and pedestrians at three difficulty levels. We split the training set of 7,481 images randomly into training and validation datasets in a ratio of 8 to 2. The accuracy of the model regarding the average precision (AP) is evaluated on the validation dataset. 

For the training, the batch size was 64 and the initial learning rates of FRDet was set at 0.0005.
The training was conducted on an NVIDIA GTX1080Ti with CUDA 10.0 and cuDNNv8. The validation of the model is conducted on an embedded GPU processor (NVIDIA Xavier) with Jetpack v4.4 set to the highest performance mode.
The baseline models for comparison were trained and evaluated on the KITTI dataset under the same settings. 

First, the trade-off between the AP, model size, and speed of BFLOPS by tuning the hyperparameter of the squeeze ratio was analyzed. Then, with the selected optimal squeeze ratio, the efficacy of each network module was studied. Finally, the proposed model with some variants was compared with the selected baseline models in terms of accuracy, FPS, and model size.


\begin{figure}[t]
\begin{center}
   \includegraphics[width=0.95\linewidth]{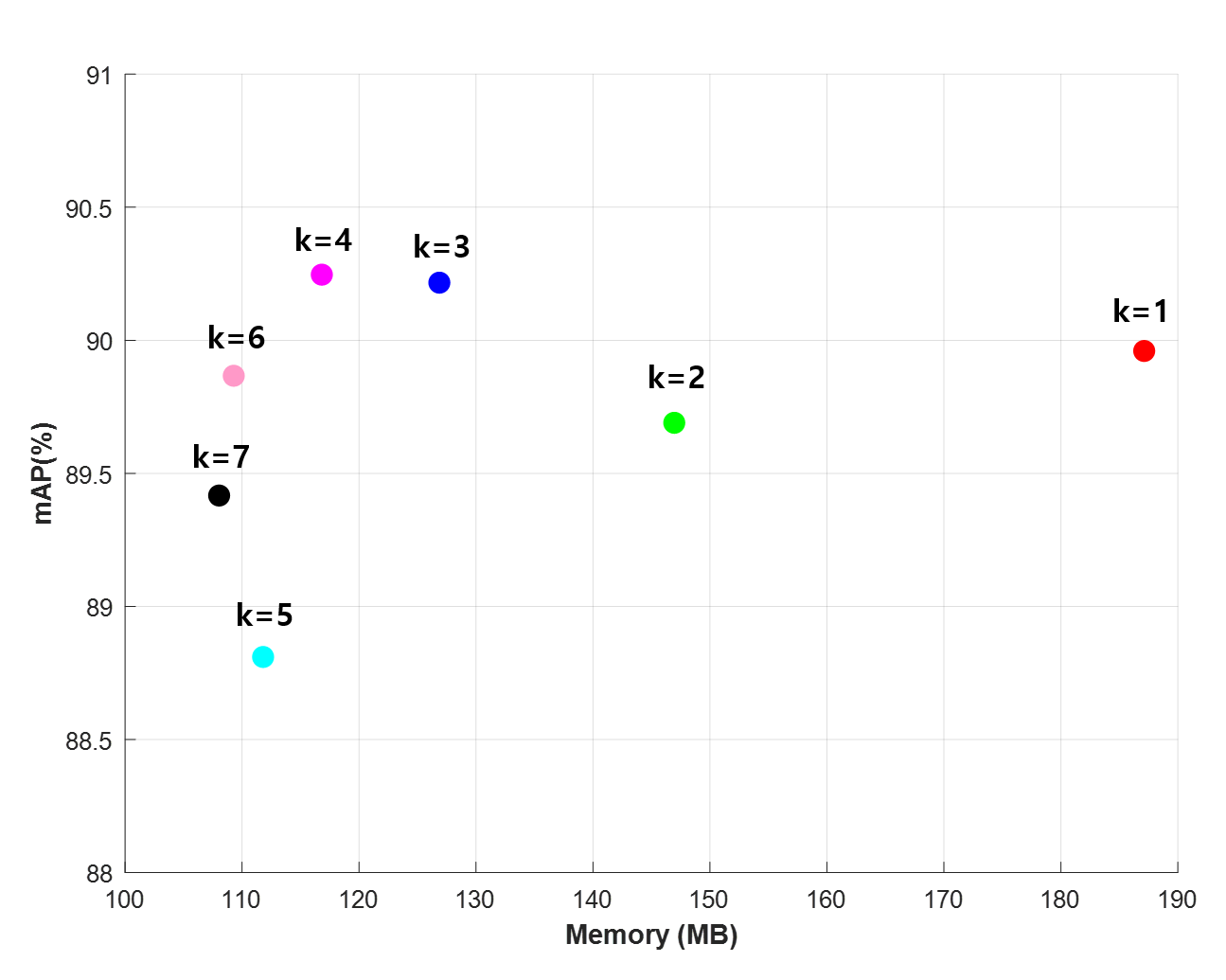}
\end{center}
   \caption{The trend of accuracy and memory size by the squeeze ratio ${2^k}$. Experiment was conducted with KITTI dataset.}
\label{fig:squeeze_ratio_graph}
\end{figure}


\begin{table}
\begin{center}
\begin{tabular}{ >{\centering}m{.2\linewidth}  >{\centering}m{.17\linewidth}  >{\centering}m{.2\linewidth} >{\centering}m{.17\linewidth} } 
\hthickline
Squeeze ratio ($2^k$) & mAP [\%] & Model Size [MB] & BFLOPS \tabularnewline
\hline
$k=1$   & 89.96 & 187.12 & 48.52 \tabularnewline
$k=2$   & 89.69 & 146.95 & 36.825\tabularnewline
$k=3$   & 90.22 & 126.87 & 30.977\tabularnewline
$k=4$  & \textbf{90.25} & 116.82 & 28.053\tabularnewline
$k=5$  & 88.81 & 111.8  & 26.591\tabularnewline
$k=6$  & 89.87 & 109.29 & 25.86 \tabularnewline
$k=7$ & 89.42 & 108.04 & 25.494\tabularnewline
\hthickline
\end{tabular}
\end{center}
\caption{The effect on the overall accuracy, model size and BFLOPS by tuning the squeeze ratio ${2^k}$. The accuracy was the highest when $k=4$.}
\label{SR_map}
\end{table}

\begin{table}
\begin{center}
\begin{tabular}{ >{\centering}m{.45\linewidth}  >{\centering}m{.10\linewidth}  >{\centering}m{.10\linewidth} >{\centering}m{.10\linewidth} } 
\hthickline
Component & \multicolumn{3}{c}{FRDet16} \tabularnewline
\hline
Fire module   & \checkmark  & \checkmark & \checkmark \tabularnewline
Residual connection &  & \checkmark  & \checkmark \tabularnewline
Gaussian detection layer &  &  & \checkmark \tabularnewline
\hline
mAP [\%] & 86.31  & 90.25 & \textbf{91.26} \tabularnewline
Memory Size [MB] & 116.82 & 116.82 & 118.45 \tabularnewline
\hthickline
\end{tabular}
\end{center}
\caption{Efficacy of FRDet components: The fire module, residual connection and the Gaussian detection layer.}
\label{component_variation}
\end{table}

\begin{table*}
\begin{center}
{\footnotesize
\begin{tabular}{ >{\centering}m{3.2cm} >{\centering}m{0.5cm} >{\centering}m{0.5cm} >{\centering}m{0.5cm} >{\centering}m{0.5cm} >{\centering}m{0.5cm} >{\centering}m{0.5cm} >{\centering}m{0.5cm} >{\centering}m{0.5cm} >{\centering}m{0.5cm} >{\centering}m{1.3cm} >{\centering}m{1.4cm} >{\centering}m{1.2cm} } 

\hthickline
\multirow{2}{*}{\shortstack{Detection \\Algorithm}} & \multicolumn{3}{c}{Car} & \multicolumn{3}{c}{Cyclist} & \multicolumn{3}{c}{Pedestrian} & \multirow{2}{*}{mAP [\%]} & \multirow{2}{*}{\shortstack{Model Size\\{[MB]}}} & \multirow{2}{*}{BFLOPS} \tabularnewline
& E & M & H & E & M & H & E & M & H & & & \tabularnewline
\hline
~\textit{deep CNN model:}\tabularnewline
Gaussian YOLOv3(416)~\cite{choi2019gaussian}    & 99.46 & 99.35 & 94.37 & 96.86 & 94.06 & 91.25 & 92.08 & 86.58 & 81.45 & ~\textbf{92.83} & 240.76 & 65.384\tabularnewline
YOLOv3(416)~\cite{redmon2018yolov3}           & 99.61 & 97.12 & 91.92 & 94.4  & 90.2  & 87.75 & 88.81 & 83.09 & 78.44 & 90.15 & 240.68 & 63.355\tabularnewline
\hline\hline
~\textit{ours:}\tabularnewline
FRDet8*& 98.68 & 96.39 & 88.92 & 96.21 & 90.89 & 90.2  & 88.64 & 83.55 & 78.47 & 90.21 & 126.87 & 30.977\tabularnewline
FRDet16*& 96.4  & 96.3  & 88.86 & 93.81 & 93.45 & 90.85 & 89.1  & 84.24 & 79.21 & 90.25 & \textbf{116.82} & \textbf{28.053}\tabularnewline
FRDet8 & 99.84 & 99.32 & 94.18 & 95.74 & 90.92 & 87.93 & 89.44 & 84.33 & 79.43 & 91.24 & 126.95 & 31.006\tabularnewline
FRDet16 & 99.61 & 99.10 & 94.09 & 93.73 & 93.14 & 90.09 & 88.79 & 83.93 & 78.90 & \textbf{91.26} & 118.45 & 28.215 \tabularnewline
\hthickline
\end{tabular}
}
\end{center}
\caption{Performance comparison with deep CNN based detectors using the KITTI validation set. E, M, and H refer to easy, moderate, and hard, respectively. FRDet* models refer to exclusion of Gaussian detection parameters.}
\label{comparison_table}
\end{table*}

\begin{table}
\begin{center}
\begin{tabular}{ >{\centering}m{.48\linewidth} >{\centering}m{.11\linewidth}  >{\centering}m{.13\linewidth} >{\centering}m{.11\linewidth}} 
\hthickline
Detection Algorithm & mAP [\%] & FPS on Xavier & Memory [MB] \tabularnewline
\hline
~\textit{deep CNN model:}\tabularnewline
Gaussian YOLOv3(416)~\cite{choi2019gaussian} & 92.83 & 23.5 & 240.76\tabularnewline
YOLOv3(416)~\cite{redmon2018yolov3}  & 90.15 & 23.6 & 240.68\tabularnewline
\hline\hline
~\textit{compressed:}\tabularnewline
SqueezeDet~\cite{wu2017squeezedet}  & 84.72 & - & 8.3 \tabularnewline
Tinier YOLO~\cite{fang2019tinier}  & 63.05 & 142.7 & 8.3\tabularnewline
SSDLite(224)~\cite{sandler2018mobilenetv2}  & 36.58 & 34.0 & 12.76\tabularnewline
\hline\hline
~\textit{ours:}\tabularnewline
FRDet8* & 90.21 & 30.3 & 126.87\tabularnewline
FRDet16* & 90.25 & 31.5 & 116.82\tabularnewline
FRDet8  & 91.24 & 30.1 & 126.95\tabularnewline
FRDet16  & 91.26 & 31.3 & 118.45\tabularnewline
\hthickline
\end{tabular}
\end{center}
\caption{Performance comparison results on embedded GPU(NVIDIA Xavier), with video input size of 480x360.}
\label{fps_xavier}
\end{table}

\subsection{Ablation studies} \label{ablation_study}
\vspace{0.3\baselineskip}
\textbf{Squeeze Ratio:} 
The memory size of the detector depends heavily on the squeeze ratio in the FR module. The parameter number of the FR module can be calculated by the numbers of 1x1 or 3x3 kernels, as indicated in Equation 3. In particular, the value $s_{1\times1}$ of the squeeze layer is the key factor in reducing the FR module size.
The squeeze ratio is defined as ${2^k}$ from the relationship of $s_{1\times1, k} = \frac{C}{2^k}$, as described in the FR module design strategy of Section ~\ref{FRmodule_description}.

The effect on the model size, BFLOPS, and the overall accuracy by tuning the squeeze ratio from k=1 to 7 is studied. As shown in Table~\ref{SR_map}, the accuracy tendency has no regular pattern but exhibits relatively superior performance when $k$ is three or four
Also, the memory size tends to decay exponentially with the increase in the squeeze ratio as shown in Figure~\ref{fig:squeeze_ratio_graph}.

When the squeeze ratio factor, $k$, is outside the range of three and four, the accuracy of the detector model tends to degrade. 
With a large squeeze ratio for a smaller number of 1D convolution kernels ($s_{1\times1}$) in the squeeze layer, the compressed feature map size might not be sufficient to extract key features. Conversely, when the number of $s_{1\times1}$ is much larger with a low squeeze ratio factor, it may have adverse effects such as overfitting. Considering the accuracy and the model size reduction, we chose $k=4$ as the optimal value of squeeze ratio.

\textbf{Components of FRDet:} 
With the squeeze ratio set at $k=4$, an experiment was conducted to explore the efficiency of each FRDet component regarding accuracy and memory size. The key components of the FR module are the fire module(without the residual connection), residual skip connections, and the Gaussian-parameter-based detection layer. 
The results in Table \ref{component_variation} show that adopting the residual skip connection to the fire module improved the accuracy by mAP 3.94\% without additional memory gain. Moreover, the accuracy was further improved by mAP 1\% when the Gaussian parameter was used in the detector layer with only a marginal increase in the memory size. 
Thus, the experimental results validated the adoption of the residual skip connection and the Gaussian detection layer to enhance the detection accuracy while maintaining memory efficiency.

\begin{figure*} 
\begin{center}
\includegraphics[width=0.9\linewidth]{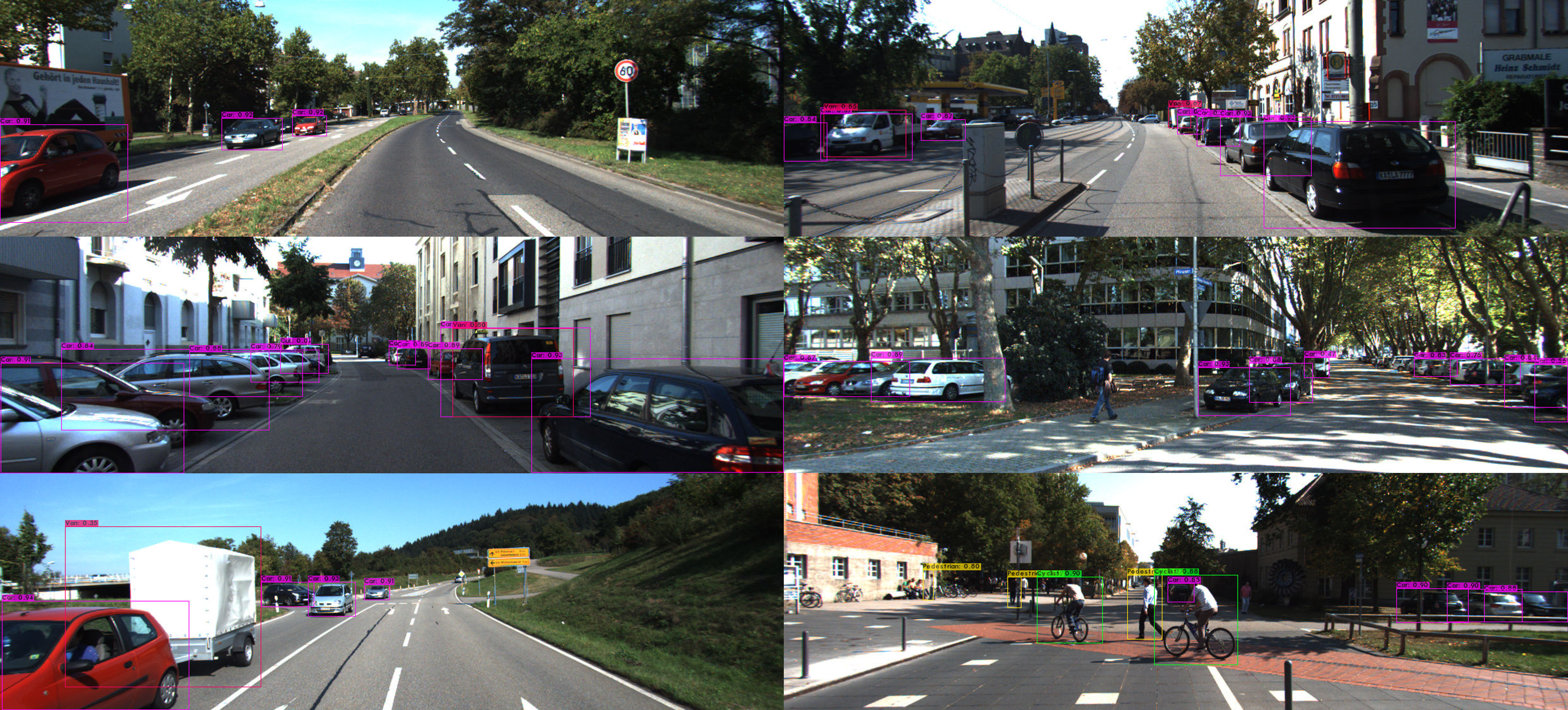}
\end{center}
\caption{Examples of detection results by the proposed FRDet16 on the KITTI dataset. Each color box represents a different object class.}
\label{fig:detection_results}
\end{figure*} 

\subsection{Performance evaluation of FRDet}

The proposed FRDet was compared to selected baseline models for performance evaluation. The baseline models were grouped into two categories: (1) deep CNN-based real-time object detection and (2) lightweight object detection for mobile processors.
Because the FRDet is based on YOLOv3 architecture, the first group of baseline models was comprised of YOLOv3~\cite{ redmon2018yolov3} and Gaussian YOLOv3\cite{choi2019gaussian} units.
For the second comparison group, high-performance lightweight object detectors based on YOLO and the fire module, which is similar to our model, were chosen with Tinier-YOLO\cite{fang2019tinier} and SqueezeDet\cite{wu2017squeezedet}. Also, SSDLite\cite{sandler2018mobilenetv2}, which is a widely used lightweight model for embedded GPUs, also is compared. 

The baseline models were trained and tested on the same KITTI dataset and settings for a fair comparison. In addition, their default hyperparameter settings were used. 
The performance evaluation was compared in terms of mAP (mean average precision), BFLOPS (billion floating-point operations per second), and memory size on an NVIDIA Jetson Xavier GPU to verify its applicability on an embedded processor.

The evaluation metrics for detecting all classes of cars, cyclists, and pedestrians are compared in Table \ref{comparison_table} and Table \ref{fps_xavier}. Compared to YOLOv3 and Gaussian YOLOv3, the memory size, and FLOPS of FRDet were reduced by approximately 50\% as indicated in Table \ref{comparison_table}. The average accuracy of the proposed Gaussian FRDet16 surpassed YOLOv3 even with only 50\% of the model’s size. The average accuracy is approximately 1.6\% lower than that of Gaussian YOLOv3. However, considering the model size and FLOPS reduction, FRDet is merited for efficient real-time object detection with high accuracy. 

As shown in Table~\ref{fps_xavier}, the lightweight models such as Tinier-YOLO, SqueezeDet, and SSDLite were heavily weighted towards model size reduction at the cost of accuracy. Because FRDet’s balance is weighted toward both reduction and accuracy, the model’s size is comparatively much larger. However, the accuracy of FRDet is 6.5-54\% higher than that of the lightweight models.

To verify the performance of the FRDet on an embedded platform, we measured the detection speed (FPS) on NVIDIA Xavier, the embedded GPU processor. For the experiment, each model was implemented on the processor and tested on a video 480$\times$360 in size. The inference speed (FPS) on the NVIDIA Xavier board was compared with the baseline models, as summarized in Table~\ref{fps_xavier}. FRDet is measured to detect in real-time with a speed of 31.3 FPS, which is about 33\% faster than YOLOv3.

In summary, compared to lightweight object detectors, FRDet has a higher model size but is advantageous with superior accuracy. Compared to deep CNN object detectors, FRDet has comparable or higher accuracy even when the model’s size is reduced by half. 
Clearly, FRDet has the advantage of being in the category of an object detector with a more balanced trade-off between the model size and accuracy while maintaining a real-time inference on an embedded GPU board. 

\section{Conclusion}
In this paper, we proposed FRDet, a lightweight object detector that is well balanced regarding the accuracy, memory size, and real-time inference on an embedded processor for practical application in autonomous driving. 
We designed the FR module that adopts fire modules with residual skip connections to compress the model size with little loss of accuracy. Moreover, the bounding box is modeled by Gaussian parameters further to enhance the accuracy of localization. 
The proposed network achieved a much higher memory reduction than conventional deep CNN-based object detectors. Moreover, the accuracy of FRDet achieved a higher value than the lightweight object-detector baselines and even surpassed YOLOv3. 
In comparison to YOLOv3, FRDet has a memory reduction of 50.8\% but achieves higher accuracy by 1.12\% on the KITTI dataset. Moreover, the inference speed was 31.3 FPS on the embedded NVIDIA Xavier GPU.
Therefore, the proposed FRDet is a well-balanced and efficient object detector for practical application to autonomous driving that can satisfy all the constraints of accuracy, model size, and real time processing on an embedded processor.

{\small
\bibliographystyle{ieee_fullname}
\bibliography{Arxiv_FRDet}
}

\end{document}